\pdfoutput=1

\documentclass[11pt]{article}
\usepackage{setspace}
\usepackage[dvipsnames]{xcolor}
\usepackage{graphicx}
\usepackage{tabularx}
\usepackage{afterpage}
\usepackage{booktabs}
\usepackage{amssymb}
\newcommand{\ocircle}{\textcircled{\raisebox{-0.9pt}{}}}
\usepackage[]{acl}

\usepackage{times}
\usepackage{latexsym}
\usepackage{xspace}
\usepackage[T1]{fontenc}
\usepackage[utf8]{inputenc}

\usepackage{microtype}

\usepackage{inconsolata}
\usepackage{float}

\title{How did we get here? Summarizing conversation dynamics}

\author{Yilun Hua,  Nicholas Chernogor, Yuzhe Gu \\ {\bf  Seoyeon Julie Jeong,  Miranda Luo, Cristian Danescu-Niculescu-Mizil}\\
Cornell University \hspace{8pt} University of Pennsylvania \\
\texttt{yilunhua@cs.cornell.edu} \hspace{8pt} \texttt{tracygu@seas.upenn.edu} \\
\texttt{\{nac86, sj598, mml267\}@cornell.edu} \hspace{8pt} \texttt{cristian@cs.cornell.edu
}
}
\newcommand{\xhdr}[1]{{\noindent\bfseries #1.}}

\newcommand{\subtext}{dynamics\xspace}
\newcommand{\dynamics}{dynamics\xspace}

\newcommand{\gptelaborate}{procedural prompt\xspace}

\newcommand{\informativeness}{informativeness\xspace}
\newcommand{\tone}{tone\xspace}
\newcommand{\Tone}{Tone\xspace}
\newcommand{\strategies}{conversation strategies\xspace}
\newcommand{\Strategies}{Conversation strategies\xspace}
\newcommand{\strategy}{conversation strategy\xspace}

\newcommand{\SCDs}{SCDs\xspace}
\newcommand{\SCD}{SCD\xspace}

\begin{document}

\maketitle

\begin{abstract}

Throughout a conversation, the way participants interact with each other is in constant flux: their tones may change, they may resort to different strategies to convey their points, or they might alter their interaction patterns.
An understanding of these \emph{\dynamics} can complement that of the actual facts and opinions discussed, offering a more holistic view of the trajectory of the conversation:
how it arrived at its current state and where it is likely heading.

In this work, we introduce the task of summarizing the \dynamics of conversations, by constructing a dataset of human-written summaries, and exploring several automated baselines.
We evaluate whether such summaries can capture the trajectory of conversations via an established downstream task: 
forecasting whether an ongoing conversation will eventually derail into toxic behavior.
We show that they help both humans and automated systems with this forecasting task.
Humans make predictions three times faster, and with greater confidence, when reading the summaries than when reading the transcripts.
Furthermore, automated forecasting systems are more accurate when constructing, and then predicting based on, summaries of conversation dynamics, compared to directly predicting on the transcripts.

\end{abstract}

\section{Introduction}
\label{sec:intro}

Conversations take place on at least two different levels \cite{tannen_conversational_2005}. 
On one level, participants directly communicate ideas, facts, and opinions, providing the topical context of the discussion.
On the other level, the \textit{dynamics} of their interactions reveal how they feel about each other, through changes in their tone---e.g., polite \cite{lakoff_logic_1973,brown_politeness_1987}, condescending \cite{huckin_critical_2002}, or sarcastic \cite{jorgensen_functions_1996}---, conversational strategies they employ---e.g., rhetorical questions \cite{han_interpreting_2002}---and even the patterns of their exchanges \cite{sacks_simplest_1974,silverstein_pragmatic_1984}.

\begin{figure}[h!]
    \centering
    \fbox{
          \begin{minipage}{\dimexpr\textwidth/2-0.9cm\relax}%
           \small
            \textbf{Traditional summary:} 
            In this online conversation, participants discuss capitalism and its sustainability. Speaker1 argues that profit always trumps morals and ethics in business. Speaker2 disagrees, stating that unchecked capitalism is economically unsustainable and detrimental to human dignity. Speaker4 seeks clarification on the consequences of capitalism's unsustainability, and Speaker2 suggests it could lead to economic collapse, mass unemployment, and toxic environments. They emphasize the need for checks on capitalism to prevent these issues. The conversation highlights concerns about unregulated capitalism and its impact on society.
            \end{minipage}
    }
    \fbox{
         \begin{minipage}{\dimexpr\textwidth/2-0.8cm\relax}%
            \small
            \textbf{Summary of conversation dynamics (human-written):}
            Several users discuss regulation of capitalism. Speaker2 shares an opinion which Speaker4 questions. Speaker2 addresses Speaker4’s question \textcolor{blue}{in a sincere manner.} Then, the \textcolor{blue}{overall tone shifts to aggressive and confrontational} when Speaker4 \textcolor{blue}{rudely invalidates} Speaker2’s response. Speaker2 \textcolor{blue}{sarcastically criticizes} Speaker4’s attitude which aggravates Speaker4 more. Then, Speaker2 \textcolor{blue}{genuinely apologizes and elaborates} on their point and Speaker4 \textcolor{blue}{gratefully acknowledges} this and clarifies their intentions. They \textcolor{blue}{continue the discussion in a calmed down, civil tone.}
           \end{minipage}}
    \fbox{%
         \begin{minipage}{\dimexpr\textwidth/2-0.7cm\relax}%
            \small
\textbf{Summary of conversation dynamics (generated):} Four speakers engage in a discussion about the nature of capitalism and its consequences. Speaker2 expresses concerns about unchecked capitalism and argues for the need to consider human dignity. Speaker4 challenges Speaker2 to explain the consequences of an economically unsustainable capitalist system. The conversation \textcolor{blue}{becomes heated,} with Speaker2 perceiving Speaker4's questions \textcolor{blue}{as confrontational.} Speaker2 defends their views and \textcolor{blue}{provides examples to support their argument.} The \textcolor{blue}{overall tone of the conversation remains argumentative, but civil.}      \
\end{minipage}}
    \caption{Traditional and \subtext summaries for the same conversation (transcript in Appendix \ref{appendix:miscellaneous}). Elements of conversational \dynamics are colored in blue.} \label{fig:text_contrast}
\end{figure}

A holistic description of a conversation and its trajectory requires accounting for both of these communication levels.  
We complement prior work that has largely focused on summarizing the topical context of the discussion \cite{yang_summarization_2023},  by introducing the task of generating summaries that instead capture the  \dynamics of the interaction between the participants.
As shown in Figure \ref{fig:text_contrast}, these cover aspects lost in a traditional summary.

Summaries of conversation \dynamics (or \SCDs for short) provide a way for humans to quickly understand the trajectory of a discussion: what type of interactions led to its current state, and how are these likely to develop?
This type of understanding can benefit various applications,
including supervision of conversations in time-sensitive domains (e.g., online community moderation, 
 supervision of mental health crisis counseling),
providing context to users (re)joining an online conversation, 
contextualizing moderator decisions,
or identifying and reviewing common problems in human-human or human-AI conversations. 
We  further discuss possible applications in Section \ref{sec:discussion} and challenges to making them feasible in Section~\ref{sec:limitations}.

However, generating \SCDs that effectively capture conversation trajectories presents several new challenges. 
While prior computational work introduced models for separately capturing individual aspects of conversation dynamics (Section \ref{sec:related}), an effective and concise summary must \emph{select} those that are most relevant for understanding the trajectory of the conversation.
Additionally, an informative summary must not simply identify these aspects separately, but should also describe how they evolve and interrelate throughout a conversation:
for example, a conversation that transitions from an aggressive tone to a calmer one has a completely different trajectory than one that proceeds in reverse order.
Thus, to provide an understanding of the trajectory of a conversation, an \SCD must \emph{synthesize} different aspects of its dynamics across multiple utterances and participants.

As a first step, we devise a multi-step procedure for human annotators to collaboratively write \SCDs.  
Importantly, this procedure is designed to address the selection and synthesis challenges described above. 
Building on this procedure, we develop a large language model prompt for generating \SCDs and compare them with summaries generated by other baselines, including traditional summaries.

Specifically, in this paper we evaluate the usefulness of \SCDs for conversation trajectory understanding via an established task: forecasting whether an ongoing conversation will eventually derail into toxic behavior \citep{zhang_conversations_2018, liu_forecasting_2018}.  
While prior attempts at this task started directly from the transcript (Section \ref{sec:related}), we explore generating \SCDs as an intermediate step.
This approach has the potential advantage of adding interpretability to automated forecasting systems and improving efficiency for humans (such as moderators) that need to make such judgments \cite{schluger_proactive_2022}.

Our findings reveal the potential of \SCDs to help both humans and automated systems understand a conversation's trajectory, motivating further work on this new task. 
In the downstream task of forecasting the future derailment of a conversation, humans make predictions three times faster, and with greater confidence, when reading the \SCD than when reading the transcript.
Furthermore, automated systems are more accurate when constructing, and then predicting based on, \SCDs compared to systems that base their forecast directly on the transcripts.
Finally, by comparing human-written and machine-generated summaries, we reveal a quality gap that motivates further computational work on this new task. 

In summary, this work:
\begin{enumerate}
    \item introduces the task of summarizing conversation \dynamics, together with a collection of human-written summaries;
    \item proposes a downstream evaluation method that allows for comparison between methods for generating \SCDs;
    \item shows the usefulness of \SCDs, motivating further work on this new task.
\end{enumerate}

To encourage future work, we release a dataset of SCDs---both human-written and machine-generated---together with the conversations they summarize in the ConvoKit library \citep{chang_convokit_2020}.\footnote{ConvoKit library: \url{https://convokit.cornell.edu/} Code and info: \url{https://github.com/CornellNLP/scd}}
This data can also support the study of individual aspects captured by these summaries, such as tone or patterns of interaction (identified via a qualitative analysis of the summaries; Section~\ref{sec:eval}).

\section{Human-written Summaries}
\label{sec:human}

To start, we introduce a procedure for writing \SCDs and a collection of such summaries for an existing dataset of online conversations.

\xhdr{Procedure for writing summaries} To construct the first collection of \SCDs, we iteratively designed a writing procedure that addresses the selection and synthesis challenges described in the introduction.
In early iterations in which we asked a single annotator to both read the transcript and write its \SCD, we observed that they consistently omitted key information that they take for granted, perhaps because some aspects of the dynamics are often processed non-consciously \cite{tannen_conversational_2005}.
To address this issue, we devise a procedure that uses interaction between two annotators to surface key elements of the conversation \dynamics that readers \textit{who cannot see the transcript} would consider relevant.
Thus, we settle on a procedure that has two parts---one in which an annotator works individually and one in which they interact with another annotator---which we briefly outline here (and detail in Appendix \ref{appendix:annotation}).

For the individual work, Annotator A will draft several summaries for a transcript in 4 steps: 
\begin{enumerate}
    \item skim over the transcript to have an overview of the topic and of the role of each speaker;
    \item read the transcript utterance by utterance and write a comprehensive summary, including opinions and arguments expressed within most utterances, turning points, and elements of conversation \dynamics;
    \item condense the summary by selecting key points and aspects of the \dynamics and replacing specific opinions and arguments with high-level descriptions;
    \item write a brief summary for each of the main speakers, focusing on (the changes in their) tone and on their conversational strategies.
\end{enumerate}

In the interactive part, Annotator B will write the \SCD, by interacting with Annotator A with a goal of understanding the conversation trajectory. 
In this process, Annotator B may read the summaries written in the previous steps by Annotator A, but not the transcript, and may make inquiries on details they deem important to understand the trajectory, such as `was this said neutrally, or is there something about the tone that I should note?' or `is the comment overtly rude, or is it just passive-aggressive or blunt?', surfacing key aspects that were not explicitly mentioned in Annotator A's summaries. 

\xhdr{Conversation transcripts data} 
\label{transcript_data}
We apply this procedure to summarize conversations from the Conversations Gone Awry - ChangeMyView (CGA-CMV) dataset \citep{chang_trouble_2019},\footnote{Accessed via the ConvoKit library.} a conversation corpus collected from the `Change My View' subreddit, where people actively seek to have others challenge their views on controversial topics.
This community has been studied extensively in part because of their explicit norms against toxic behavior, and corresponding labels inferred from the moderators' interventions. 

In the CGA-CMV corpus, conversations are paired such that every conversation that derailed---i.e., ended in a toxic comment removed by moderators---is matched with another conversation on the same topic that did not.
For us, these labels provide an opportunity to test the extent to which \SCDs provide an intuition about the future trajectory of the conversation (i.e., will it derail or not). 
To focus on the \textit{future} trajectory, we use \textit{truncated} transcripts obtained by removing the last 3 utterances from every conversation (in addition to the toxic comment, if there was one).
Since our interest is in summarization, we only keep pairs where both conversations are longer than 10 utterances. 

\xhdr{Collection of human-written summaries} 
We produce human-written summaries for 50 conversations from the train split of CGA-CMV.
The summary writing process took roughly 140 annotator-hours.\footnote{For each conversation transcript, the individual part takes about 2 hours and the interactive part takes about 20 minutes.}
Summaries are on average 71 words long (annotators are instructed to keep them under 80); for comparison, the transcripts are on average 940 words long.
An example summary is shown in  Figure \ref{fig:text_contrast}, and a qualitative analysis is provided in Section \ref{sec:eval}.

\xhdr{Informativeness check}
Before we proceed, we check whether the summaries are actually informative.  
Given their highly abstractive nature, there is a risk that they become so general as to not distinguish between different conversations (e.g., `Speaker1 disagreed with Speaker2.' would apply to most of the conversations in the data).
Thus it is not sufficient to judge whether statements in a summary are technically matching the conversation they summarize: we need to also check whether they convey sufficient information to distinguish that conversation from others.
We devise a procedure for systematically checking whether this is the case as follows. 

We ask new annotators to read a transcript, and then present them with a multiple-choice question.
Each choice corresponds to a summary segment involving two speakers.
One of the choices is from the actual summary of the provided transcript, while the other two are distractors:
one from the summary of the paired conversation (thus, on the same topic, but with the opposite derailment label) and the other from the summary of another conversation with the same label as the transcript, but on a different topic.
This way, neither the topic nor the label fully reveals the answer:
to be identified correctly, the segment must contain information that matches the transcript better than the distractors.

For instance, for our introductory example, three choices could be: 
``SpeakerX sarcastically criticizes SpeakerY’s attitude which aggravates SpeakerY more.'' (an actual segment), 
``SpeakerX poses a rhetorical question, which SpeakerY contradicts sarcastically, raising the tension and causing SpeakerX to disagree rudely.'' (a same-pair distractor), ``SpeakerX first shares their opinion and later poses rhetorical questions, and SpeakerY disagrees in a matter-of-fact manner.'' (a same-label distractor).

Though we designed this procedure to avoid excessive workload when evaluating informativeness,\footnote{An equivalent check could be implemented by providing one summary segment and three transcripts to pick from. 
This method corresponds to the existing literature in communication-based evaluations for natural language generation \citep{newman_communication-based_2020}, and implements the idea that an informative summary should capture the salient information that makes the source text stand out with respect to other related texts \citep{zhang_learning_2018}.
However, this equivalent method would require substantially longer time due to the lengths of the transcripts. 
} each question still requires reading one transcript and carefully checking the segment choices against it. Therefore, we limit our total number of questions to 10, covering 30 conversations through distractors. (Further details in Appendix \ref{append:informativeness}.)
Two annotators completed the task. One answered 10 out of 10 questions correctly and the other answered 8 of them correctly (noting low confidence on the 2 answers they got wrong), suggesting that our summaries indeed pass this basic informativeness check. 

\section{Machine-generated Summaries}
\label{sec:automatic}

We now turn to explore several simple baselines for generating \SCDs, setting the stage for developing more specialized methods in future work.
Models in the GPT family have achieved remarkable results in various summarization benchmarks \citep{zhang_benchmarking_2024, yang_exploring_2023}. Among them, ChatGPT is particularly suitable for adapting to new tasks like ours without demanding a sizable train set. 
Thus, for the first group of baselines, we query OpenAI's ChatGPT (GPT-3.5-turbo-0613) API with default parameters using different prompts, from the most common prompt for traditional summarization tasks to prompts inspired by the procedure we developed for humans:\footnote{We prompt the model to generate summaries of at most 80 words and set the max new token limit to 128 (corresponding to approximately 96 words) as a hard limit .}

\xhdr{Traditional prompt} After experimenting with several prompts on a development set, we use a concise prompt for our traditional summarization baseline: `briefly summarize the following online conversation in 80 words.' Figure \ref{fig:text_contrast} includes a traditional summary generated by this prompt.

\xhdr{Zeroshot prompt} 
We devise a prompt that explicitly integrates our goal of generating summaries that can help people understand the conversation trajectory. 
After experimenting with several word choices for referring to trajectory, \subtext, and specific \subtext elements, we settle on a concise prompt, `write a short summary capturing the trajectory of the online conversation’ with additional constraints such as 
excluding specific arguments and capturing elements of tone and conversation strategies (Figure \ref{fig:prompt_contrast} in the Appendix).

\xhdr{Procedural prompt} We build on the insights we gathered from developing the procedure for human annotators (Section \ref{sec:human}) to construct a more elaborate prompt.
This prompt (Figure \ref{fig:prompt_contrast} in the Appendix) thus includes instructions adapted from those provided to the annotators, together with examples that they found useful for understanding the instructions.
Because we only include segments of summary examples instead of complete transcript and summary pairs, the procedural prompt can be positioned in-between zeroshot and few-shot in-context learning. Figure \ref{fig:text_contrast} shows the procedural prompt summary for our introductory example.

We also experimented with few-shot in-context learning on a small subset of the training set, but manual inspection did not reveal an increase in quality.  
Thus, due to significantly higher API costs, we did not pursue this path. 
Appendix \ref{appendix:autosumm} includes more discussion of our prompt engineering.

\xhdr{Finetuning} Finally, we experimented with finetuning GPT-3.5-turbo and smaller dialogue summarization systems (BART-large and DialogLED) \citep{lewis_bart_2020, zhong_dialoglm_2022} using the 50 human-written summaries and the corresponding transcripts. We provide details in Appendix \ref{appendix:autosumm}. 

\section{Downstream Evaluation: \\ \ \ \ \ \ \ {Forecasting Derailment}}
\label{sec:downstream}

\label{sec:downstream_use}
Popular 
metrics for summarization---e.g., ROUGE \citep{lin_rouge_2004}, BERTScore \citep{zhang_bertscore_2020}, and QA-based metrics---are notoriously unreliable when evaluating LLM-generated summaries or summaries of long documents \citep{koh_empirical_2022,goyal_news_2023}.
We thus follow recommendations of \citet{deutsch_statistical_2021} and perform a downstream evaluation, in which we quantify the extent to which \SCDs provide an understanding of the conversation trajectory.

Specifically, we choose the task of forecasting whether a conversation will eventually derail into toxic behavior \cite{zhang_conversations_2018}. 
Unlike previous work in which the prediction was made based on a truncated transcript of the conversation (for a comprehensive discussion of prior models see Section \ref{sec:related}), here we aim to make the prediction directly on the \SCD of that truncated transcript.  
In addition to providing means to evaluate and compare current and future models for generating \SCDs, this derailment forecasting task is also important in itself, as it was shown to enable important practical applications: automated forecasts can be used to inform users during ongoing discussions \cite{chang_thread_2022} while human forecasts are made by moderators in their everyday workflow \cite{schluger_proactive_2022} (see Section \ref{sec:limitations} for practical and ethical considerations of real-world deployment).
Besides the potential practical importance of this task, it is also worth noting its difficulty. 
Because the forecaster never actually sees the toxic comment, it must rely on subtle signals and overall trajectory of the conversation dynamics. 
This inherent difficulty and its implications on the design of forecasting models  were extensively discussed in \citet{chang_trouble_2019}. 

We first compare the usefulness of \SCDs for automated forecasting systems. 
Then we devise an experiment to estimate their usefulness for human forecasts.
Throughout, the forecasts are done on a balanced dataset of derailing and non-derailing conversations paired by topic, following the setup of the CGA-CMV dataset (Section \ref{sec:human}); thus the overall topic of the discussion plays a minimal role and the random baseline is 50\%.
To leave room for future work
we leave the original CGA-CMV test set untouched. 
%
Using truncated transcripts from the original train split, we construct a new train set (234 conversations), a new development set (100), and a new test set (100); the new test set includes the 50 conversations for which we also have human-written summaries (Section~\ref{sec:human}).

%
%

\label{automatic_pred}

\subsection{Useful for automated forecasts?}
\label{subsection:forecaster}
We train classifiers to predict if a conversation will eventually derail based on the various types of summaries of the truncated transcripts.
We adopt GPT-3.5-turbo to develop few-shot classifiers for each summary type, 
using examples from outside the test split.  
To provide more robust estimates, for each summarization method we generate 4 different summaries for each conversation, and average the classifiers' performance on them (details in Appendix \ref{append:other_cls}). 
\xhdr{Comparison of summaries}
As shown in Table~\ref{table:helping_machines}, the classifier based on the \gptelaborate achieves the best accuracy, significantly outperforming the other types of summaries ($p<0.05$; throughout we use the Wilcoxon signed-rank test significance testing). 
In particular, the information conveyed by the \SCDs generated with the \gptelaborate appears to be more useful for the automatic derailment forecaster than that included in traditional summaries. Other metrics (Macro-F1, precision, recall) support the same conclusion (Appendix \ref{subsection:additional_metrics}).

\begin{table}[b]
\centering
\begin{tabular}{lc}
\toprule
 \textbf{Based on...} & \textbf{Accuracy} \\ 
transcripts (CRAFT classifier) & 56.0  \\
transcripts (GPT-16k classifier)& 60.0 \vspace{0.05in}\\   
traditional prompt summaries & 58.3  (5.85)\\ 
zeroshot prompt summaries & 58.8 (6.24)\\ 
procedural prompt summaries & {67.3}$^{\ast}$ (2.63)  \\ 
\bottomrule 
\end{tabular}
\caption{Derailment forecasting results for systems based on truncated transcripts and on different types of machine-generated summaries. 
For summary-based systems, we report standard deviation across 4 summary-generation trials, and indicate with $^{\ast}$ the highest performance (p<0.05; Wilcoxon signed-rank test).
Results for the GPT-3.5-turbo few-shot classifiers are shown unless otherwise noted.
}
\label{table:helping_machines} 
\end{table}

Finetuned summarization models---finetuned on the 50 human-written 
summaries
and evaluated on the remaining of the test set---perform worse than the \gptelaborate on the same set (more details in Appendix \ref{appendix:autosumm}). 
This could be due to the relatively small collection of human-written summaries, as well as to the generic fine-tuning methodology, 
thus motivating extending the set of human-written summaries and developing fine-tuning  procedures that also integrate the forecasting objective.

\xhdr{Summary vs transcript}
For reference, we also include two baselines operating directly on the truncated transcripts. 
The first baseline, CRAFT, was introduced before the advent of the LLM era and remained 
a competitive system
for this task \citep{chang_trouble_2019}.\footnote{For a fair comparison, we modify the ConvoKit implementation of CRAFT \cite{chang_convokit_2020} to trigger forecasts exactly 3 utterances before the end of the conversation (in the original setup the system could make predictions all the way up to right before the attack or the end of the conversation).  
Empirically, this setup turns out to be harder for CRAFT than the original one.
While we use CRAFT as our non-LLM baseline because it relies only on the text of the conversation, we note that recent work showed that integrating user dynamics and up/down votes can lead to better performance on the CMV section of the CGA dataset \cite{altarawneh_conversation_2023}. 
Future work could explore the potential benefits of factoring in such extra-textual information into the creation of the \SCDs.
}
The second baseline is a few-shot GPT-3.5-turbo-16k classifier, which can take up to 16k tokens to cope with the greater input lengths of the transcripts.\footnote{Both baseline systems might have an advantage in that they might have accessed the full untruncated transcripts during pre-training.} 

As shown by Table \ref{table:helping_machines}, predictions based on \gptelaborate outperform those based directly on the transcripts.  
This suggests that \SCDs are effective in distilling from the transcripts information that is useful for the forecasting task.
Perhaps more importantly, the feasibility of this `summarize-then-forecast' approach points out a promising future direction for improving the interpretability of the user-facing forecasting systems, where the summary could be presented as an easily digestible rationale for the prediction. 
In fact, users of such systems have identified the lack of explanations as one of their most important drawbacks \cite{chang_thread_2022}. 

\xhdr{Other forecasting systems} We also train BART \cite{lewis_bart_2020} and longformer (LF) \citep{beltagy_longformer_2020} as finetuned classifiers for the forecasting task (Appendix \ref{append:other_cls}). While their performance is substantially lower than that of the GPT few-shoot classifier across all types of summaries, the comparisons discussed above still hold (Table~\ref{table:compare_cls}).  

\begin{table}[t!]
\centering
\begin{tabular}{@{}llll@{}}
\toprule
\textbf{Based on...}         & \multicolumn{3}{c}{\textbf{Acc. by forecaster}} \\ 
                   & GPT & BART      & LF   \\  %
transcripts         &   60.0 &     50.0       &     51.0                \\
traditional summaries     &       58.3 &    56.0        &      58.3           \\
procedural summaries      &     \textbf{67.3}  &   \textbf{63.0}   &   \textbf{61.5}           \\ 
\bottomrule
\end{tabular}
\caption{Comparing different classifier architectures for derailment forecasting. ``GPT'' refers to the GPT-3.5 few-shot classifier for summaries and to GPT-16k classifier for transcripts (same as in Table~\ref{table:helping_machines}).}
\label{table:compare_cls}
\end{table}

\subsection{Useful for human forecasts?}
\label{subsection:human_eval}
We now switch to the other main motivation: can \SCDs help \textit{humans} quickly grasp the trajectory of a conversation?
To answer this question we devise an experiment in which subjects are asked to guess whether a conversation will eventually derail based either on a transcript or its \SCD. 
We compare both their accuracy and efficiency, in terms of the time they spend to make their guess, as well as their confidence in their guess.

To better focus our resources, we use a subset of 20 paired conversations out of those for which we created human summaries.    
In addition to the transcripts and the human-written summaries, we also consider the corresponding \gptelaborate summaries (since those were shown to fare best in the automatic prediction task).

We recruit 20 university students fluent in English as participants. 
A subset of participants make their guesses based on the transcripts only, while another subset make guesses based on summaries only.
Each participant in the latter subset sees a mix of human-written and machine-generated summaries (without being aware that these are produced differently) such that any observed differences between their effects cannot be attributed to participant idiosyncrasies. 
In addition to providing a guess of whether the conversation will derail or not, each participant is asked to 
rate their confidence in their guess (on a scale from 1 to 5).
We also record the time it took for the participants to make their guess (starting from the time they see the transcript or summary until the time they select their guess), and instruct them to work on each question without pausing. 
The specific instructions and details about how participants are grouped are in Appendix \ref{appendix:eval}.

Unlike in the automatic evaluation in Section \ref{automatic_pred}, we adopt a zeroshot prediction setting, 
in which humans do not have labeled examples of summaries (or transcripts) to assist their guessing.
This way, we can better test if the summaries are immediately intuitive to humans rather than testing the participants' ability to learn patterns that might not be visible to untrained individuals. 
This means, however, that the accuracies of the human participants are not directly comparable with those of the automated system.

\xhdr{Summaries vs transcripts} As shown in Table~\ref{table:human_on_human_machine},  participants can make guesses 3-4 times faster based on \SCDs while maintaining similar accuracy.
This improvement in efficiency is critical for applications such as proactive online moderation, as earlier work has found that moderators are faced with ``too many [potentially at-risk conversations] to proactively monitor'' \citep{schluger_proactive_2022}.

\xhdr{Human vs generated summary}  Participants are significantly more confident when making predictions based on human-written summaries than on machine-generated summaries (and even on the transcripts).\footnote{This difference continues to hold when only considering correct guesses. Also, reassuringly, confidence in correct guesses is higher than in incorrect ones throughout.}  
This gap is important for applications where summaries are used for decision-making (e.g., moderation) and motivates future work on improving summarization models.
Another noticeable difference is that machine-generated summaries provide a better understanding of the topical content of the discussion, perhaps to the detriment of better coverage of aspects of conversation \dynamics.  
In Section \ref{sec:eval} we further explore this tradeoff via a qualitative analysis of the summaries.\footnote{We also experimented with directly asking participants to report their understanding of the trajectory of the conversation, on a scale from 1 to 5. There was no significant difference between human and machine-written summaries (4.0 and 3.9 respectively), perhaps due to the difficulty of briefly explaining what a trajectory is and how it differs from the derailment prediction, a confusion that surfaced during debriefing.}

\begin{table}[t]
\centering
\begin{tabular}{lllll}
\toprule
\textbf{Based on...} & \textbf{Acc} & \textbf{Conf}  & \textbf{Topic} & \textbf{Time} \\ 
transcripts & 60 & 3.5 &  - & 132\\ 
gen. summ.  & 59 & 3.6 & {3.9}  & 45$^{\ast}$ \\ 
human summ. & 62 & {4.0}$^{\ast\dag}$ & {3.4}$^{\dag}$ & {31}$^{\ast}$ \\ 
\bottomrule
\end{tabular}
\caption{Results on the human forecasting experiment. ``gen. summ.'' refers to the summaries generated using the procedural prompt. Time is measured in seconds. $^{\ast}$~indicates a significant difference when compared with transcripts (p<0.05; Wilcoxon signed rank test), $^{\dag}$~indicates a significant difference when comparing human-written with automated summaries (p<0.05).
}

\label{table:human_on_human_machine}
\end{table}

\section{Qualitative Analysis}
\label{sec:eval}

To complement our quantitative evaluation and understand what might drive the differences between human and machine-generated summaries, we now turn to the actual content of the \SCDs.  
Through a close reading of the 20 human-written and 20 machine-generated summaries used in the experiment described above, we identify, annotate, and compare several aspects that were shown to provide clues about the conversation trajectories. 

\xhdr{\Tone} Whether `polite,' `rude,' `aggressive,' `condescending,' or `sarcastic' \citep{brown_politeness_1987, tannen_conversational_2005}, the \tone employed by the participants is a prominent feature of the \SCDs.
Tone can be explicitly stated, as in `Speaker1 disagrees [...] \textbf{in a somewhat passive-aggressive tone}.' Other times, especially in human-written summaries, it is expressed as modifying a speech act, as in `contradicts \textbf{sarcastically}', `disagrees \textbf{rudely},' and `\textbf{adamantly} defends.'
%
%
Overall, \tone is indicated less frequently in the machine-generated summaries (75\% of them mention tone at least once) than in the human-written summaries of the same conversations (all mention tone at least once), suggesting a potential path for improvement.

\xhdr{Changes in \tone}
\Tone can evolve throughout a conversation, and changes in tone can provide an intuition about its trajectory \cite{niculae_linguistic_2015}.  
When participants use an `\textbf{increasingly} passive aggressive tone,' or when the `tension \textbf{rises}' the conversation seems more likely to be getting out of hand than when a `slight tension [...] is \textbf{maintained but doesn't escalate}' or when the `tone \textbf{remains} argumentative but civil'.  
The latter quote is an example of an overall assessment of tone dynamics that both humans and (more commonly) automated systems sometimes include at the end of the summary, even though neither is explicitly instructed to do so.  
Overall, 75\% of the human summaries feature phrases explicitly mentioning changes in \tone whereas only 50\% of the machine-generated counterparts do so.

\xhdr{Patterns of interaction} Beyond the content of the messages, the structural properties of the interactions were shown to be indicative of future trajectories \cite{backstrom_characterizing_2013, zhang_characterizing_2018}.  
Two participants can have a `\textbf{brief exchange}' or an extended `\textbf{back-and-forth}', which can be interrupted when another participant `\textbf{jumps in}'.
While explicit mentions of such patterns are relatively rare (found in 45\% of the human summaries and 31\% of the machine-generated summaries), they can often be inferred by following the sequence of speakers mentioned in the summaries.
 
\xhdr{\Strategies} Interlocutors employ strategies that can put the conversation on various trajectories.
For example, `pos[ing] a \textbf{rhetorical question}' or `\textbf{questioning each other's logic}', can often lead to personal attacks \citep{habernal_before_2018}, whereas expressing uncertainty about one's own view (e.g., via hedging), would soften an impending disagreement and prevent the escalation of tension \citep{zhang_conversations_2018}.  
`Supporting [a] point with \textbf{evidence}', `justifying objective claims with \textbf{personal experiences}', `draw[ing] a \textbf{comparison}' or `\textbf{question[ing] the importance of specific details}' are classic persuasion strategies \citep{zeng_what_2020,li_exploring_2020}. 
A list of strategies considered in this analysis is included in Appendix~\ref{appendix:qual_analysis}. 
Overall, we find that mentions of conversational strategies are similarly common in human-written (80\%) and machine-generated summaries (85\%). 

\xhdr{Topical context} Finally, these dynamics can only exist in the context of the content being discussed. 
Though not the primary focus of \SCDs, a small amount of topical context is needed to provide a scaffolding for the phenomena discussed above.
Both human and machine-generated summaries generally start with a sentence about the general topic of the discussion.
Beyond that, machine-generated summaries include substantially more topical context to the detriment of actual aspects of conversation \dynamics, despite the explicit instruction and in-context-learning examples against this behavior.
This echoes the subjective ratings of the participants in the human forecasting experiment (Table~\ref{table:human_on_human_machine}).
This phenomenon suggests that in-context learning is not sufficient to `untrain' LLMs from the traditional summary examples seen in pretraining.  
This motivates developing models that are specifically designed to select and synthesize aspects of conversation dynamics, perhaps inspired by the interactive human-writing procedure.

\section{Further Related Work}
\label{sec:related}
Our work falls at the intersection of three broad areas of NLP: studies of conversation \dynamics, summarization, and conversation forecasting.  

\xhdr{Conversation dynamics} 
We are primarily inspired by extensive computational work on modeling various aspects of conversation dynamics.   
Some studies have focused on identifying specific aspects, such as such as politeness \cite{burke_mind_2008,danescu-niculescu-mizil_computational_2013,li_studying_2020}, formality \cite{krishnan_youre_2015,pavlick_empirical_2016}, passive-aggressiveness \cite{chhaya_frustrated_2018}, condescension \cite{wang_talkdown_2019}  or sarcasm \citep{oraby_are_2017}.
Others have studied changes along these dimensions during the discussion \cite{wang_piece_2014, niculae_linguistic_2015, niculae_conversational_2016}.
A separate but related thrust focused on persuasive strategies interlocutors employ in a conversation, mostly in the context of debates (see \citet{lawrence_argument_2020} for a survey).
Unlike these studies, the goal of \SCDs is not to exhaustively identify occurrences of either one of these phenomena, but to convey how such key aspects combine towards an understanding of the conversation's trajectory.
Nevertheless, the dataset of SCDs that we release (with annotated aspects of conversation dynamics) can constitute an additional resource for studying these phenomena and the context in which they occur. 

\xhdr{Dialogue summarization} The vast majority of dialogue summarization systems focus on the content of the utterances, rather than on the more subtle non-topical \subtext.
Early approaches to dialogue summarization focused on using external tools to explicitly model dialogue structures, such as topic segmentation and conversation stages \citep{li_keep_2019, chen_multi-view_2020}, dialogue acts \citep{goo_abstractive_2018}, discourse dependency and speaker-action relations \citep{chen_structure-aware_2021}, which are processed into features that can help language models. 
Later, pretraining on dialogue corpora also attracted increasing research interest and achieved state-of-the-art results on many datasets \citep{zhong_dialoglm_2022}. 
Most recently, extensively pretrained, instruction-tuned LLMs, such as the GPT family models, have achieved superior results on various summarization leaderboards \citep{goyal_news_2023, zhang_benchmarking_2024, yang_exploring_2023}. 
In dialogue summarization,  these instruction-tuned LLMs possess strong in-context-learning capabilities \citep{wu_brief_2023}, making them strong candidates for solving new summarization tasks that have limited training data. 

\xhdr{Conversation forecasting} We motivate and evaluate \dynamics summaries with applications requiring an understanding of a conversation's trajectory.  
Beyond forecasting derailment \citep{zhang_conversations_2018, liu_forecasting_2018, chang_trouble_2019}, other tasks include forecasting thread growth \cite{backstrom_characterizing_2013}, prosocial outcomes \cite{bao_conversations_2021}, editorial decisions \citep{mayfield_analyzing_2019}, controversy \cite{hessel_somethings_2019}, the outcomes of negotiations \citep{chawla_bert_2020} or team resilience \citep{whiting_did_2019}.
It would be interesting to consider the extent to which \SCDs can aid with these other forecasting tasks, and how to obtain summaries specifically dedicated to these tasks.

To the best of our knowledge,  all conversational forecasting systems operate directly on conversation transcripts. 
The early work by \citet{chang_trouble_2019} adopts a recurrent network and applies unsupervised training to learn a representation of conversation dynamics. 
More recently, \citet{kementchedjhieva_dynamic_2021} explores pretraining and various training paradigms for this task, \citet{altarawneh_conversation_2023} applies a graph convolutional network, and \citet{yuan_conversation_2023} uses a hierarchical transformer-based framework to combine utterance-level and conversation-level information. 
However, since it aims to guess the future, this task remains challenging. 

Unlike detecting the toxic language after the fact \cite{wulczyn_ex_2017,breitfeller_finding_2019}, the signs of future derailment are subtle and require a more thorough understanding of the conversation trajectory. 
Our results suggest that \SCDs can provide this information concisely and effectively, suggesting a new summarize-then-forecast approach to conversational forecasting tasks.
This inspires future work to integrate \SCDs in real-time forecasting systems, which would require tackling shorter conversations where summaries might not be appropriate, as well as the `unknown horizon' problem: not knowing when to trigger the prediction \cite{chang_trouble_2019}.

\section{Conclusions}
\label{sec:discussion}

In this work, we introduce the task of summarizing the dynamics of interaction between participants in a text-based conversation.   
By introducing human and automated procedures for writing such summaries, we show that they can capture information that is mostly missing from traditional summaries, such as the tone in which the participants write and how it changes throughout a conversation.
Summaries of these dynamics are useful to both humans and automated systems for understanding the overall trajectory of the conversation, as shown through the downstream evaluation task of forecasting whether a conversation will eventually derail or not.  
Humans can make similarly accurate forecasts three to four times faster by starting from \SCDs than by reading the transcripts.
When compared to directly predicting on the transcripts, automated systems make better forecasts when generating \SCDs as an intermediate step.

Going beyond the gains in accuracy and efficiency, \SCDs add interpretability to the forecasting task. 
Interpretability is key to enabling applications where forecasts are used by humans for decision making.  
For example, when warning interlocutors (or moderators) that a conversation is at risk of derailing \cite{chang_thread_2022,schluger_proactive_2022}, \SCDs could provide them with insights into why that might be the case, helping them decide whether to heed the warning and how to deescalate the situation. 
This pertains to settings beyond online forums, such as customer service conversations or mental health counseling conversations, in which supervisors need to decide fast whether (and how) to intervene in conversations that seem to enter a non-desirable path.

In addition to forecasting, capturing and succinctly describing the dynamics of conversations can be useful for a series of applications, including training conversationalists (e.g., by providing them with a way to review their prior conversations), providing context to people (re)joining a conversation (e.g., to aid a therapist prepare for their next session with their client), or identifying common trajectories in human-AI conversations.
\SCDs could even aid conversational analysis researchers more efficiently explore individual conversations and reveal patterns in the intricate ways in which they develop \cite{sidnell_conversation_2011}.

\section{Limitations}
\label{sec:limitations}

This work, however, only takes the first steps towards solving and evaluating the task of generating \SCDs automatically.
In fact, we show that there is a substantial gap remaining between human-written summaries and machine-generated ones.  
Since in this work we focus on defining the task and demonstrating its feasibility, we only employ simple prompting and standard fine-tuning procedures. 
This sets the stage for the future development of more specialized models and training regimes.
These models could be more tightly integrated with the downstream task, learning to select aspects of the \dynamics that are most relevant as well as to determine the right level of abstraction.

To continue improving on \subtext generation models, more diverse automated evaluation methods are required.  
Given the highly abstractive nature of the task, traditional metrics based on token overlap or semantic similarity are not immediately applicable \cite{goyal_news_2023}.
Our \informativeness check provides an avenue for evaluation that could potentially be scaled up through automation.
Furthermore, considering other downstream applications, such as forecasting prosocial outcomes \cite{bao_conversations_2021} or how likely it is for participants to change their mind \cite{tan_winning_2016,hovy_importance_2021}, could further help evaluate the usefulness of \subtext summaries.

While the current work is restricted to summaries of text-based conversations, important \subtext can be encoded in vocal features (e.g., intonation, or pitch) or gestures (laughter, body positioning).  
A multimodal approach could enable applications that go beyond text-based conversations and provide a more holistic understanding of conversational dynamics.  

Additionally, while we tested how useful summaries are for humans in a small-scale control setting, further work could test this more comprehensively through user studies, for example by integrating these summaries into conversational assistance tools \cite{chang_thread_2022} or moderation assistance tools \cite{schluger_proactive_2022}.
From a technical perspective, a real-time deployment would require iteratively generating summaries in real-time, as the conversation progresses, rather than at a set moment in the conversation as we do in this work for the sake of scalability.

\textbf{Ethical concerns} surrounding fairness and bias should necessarily take center stage in any deployment of summarization systems, especially since \SCDs may include mentions of emotions and affect of the people involved in the conversation \cite{zhou_entity-based_2023}.
Any broad usage scenario should undergo rigorous scrutiny of potential for unintended consequences \cite{weidinger_taxonomy_2022}.
For example, \SCDs and automated forecasts relying on them should not be used to make automated censoring or moderation decisions, in order to avoid propagating biases embedded in the underlying large language models.
If future developments will result in summaries that are reliable enough to inform human decisions (e.g., helping moderators decide whether to closely monitor an ongoing conversation), the users should be informed about systematic mistakes the summary is likely to make in that respective setting.

\section*{Acknowledgements}
We would like to start by expressing our deepest gratitude to 
Jonathan P. Chang who advised us throughout, from defining the task all the way to providing excellent suggestions on the submission draft.
We thank Tushaar Gangavarapu for his help with adapting the CRAFT baseline to our setting.
We were lucky to have had many enlightening conversations about this work during Team Zissou's meetings, which included  
Jonathan P. Chang, 
Tushaar Gangavarapu, 
Dave Jung,
Lillian Lee,
Vivian Nguyen,
Tony Wang,
Sean Zhang.
We also thank 
Ido Dagan, 
Mark Johnson, and 
Alex Niculescu-Mizil for their sharing their insights with us, as well as the anonymous reviewers for their helpful suggestions.  
Lastly we acknowledge the help we got from our annotators, the participants in the NLP seminar, and the ``Conversations and Information'' class at Cornell.
This research was supported in part by an NSF CAREER award IIS-1750615, a LinkedIn Research Award, and an Oracle ERO Award.

\bibliography{refs}

\newpage\hbox{}\thispagestyle{empty}\newpage

\appendix
\section{Instructions for Writing Summaries}
\label{appendix:annotation}

In this section we explain our annotation procedure and provide definitions for the terminologies in our instructions along the way.
As described in Section~\ref{sec:human}, the procedure is divided into two parts: one in which an annotator works individually and the other in which they interact with another annotator.

\subsection{Individual Work}

Instructions for an individual annotator: 

\begin{enumerate}
    \item Depending on the \textit{complexity} of the conversations, either 1) thoroughly read the whole conversation or 2) skim through the conversation to understand the general idea
    \begin{itemize}
        \item \textit{Complexity}: number of speakers, familiarity of the topic to the annotators, length. 
For shorter conversations, it is easier to read through the whole conversation before moving on to summarizing, while for really longer ones, annotators would read a few comments at a time, summarize, read the next few, etc.
    \end{itemize}

\item Go through the conversation\textit{ comment-by-comment} and write a comprehensive summary that captures the content of each \textit{comment} and any \textit{key points}.
\begin{itemize}
    \item \textit{comment}: all speakers' utterances are in the form of reddit comments. 
    \item \textit{key points/moments}: also referred to as “turning points” are where the tension of the conversation or the speakers’ opinions notably change. Annotators should highlight them in both the original transcript and the summary in the following way: increase in tension (red), decrease in tension (blue), change in opinions towards disagreement (yellow), change in opinions towards agreement (green)
\end{itemize}
\item Then revise the comprehensive summary to 
\begin{enumerate}
    \item change any wording that’s confusing (not accurately describing the original comment)
    \item review if the summary reflects the conversation accurately (specifically the conversation dynamics and tension) and add any tone indicators that might be missing
    \begin{enumerate}
        \item Indicate changing tension (e.g. curse words, all-caps, rhetorical questions, polite words) and indicate sentiments with phrases like “sarcastically,” “passive-aggressively,” “politely,” etc. Use direct quotes (no need to explicitly describe the emotion) if they are concise and hard to capture in a summary. Focus on the highlighted elements of the conversation when adding indicators in order to capture changes in tension.
    \end{enumerate}
    \item Condense the summary to 150 words while trying to preserve the turning points from step 2 and tone indicators indicated during revision. Omit the parts of the conversation that didn’t contribute much to the overall trajectory and otherwise reword for brevity. For example,
    \begin{enumerate}
        \item Condense lengthy or redundant back-and-forth conversation that doesn’t introduce new points (but may impact tension) into fewer sentences summarizing the main developments
        \item Omit irrelevant comments (e.g. brief interjections by a new user that did not have any substantial follow-ups)
        \item Change a few direct quotes/details to more concise sentiment words (ex. ``calling this blatant racism'' → ``... with condemnation'')
        \item Other editorial changes

    \end{enumerate}

\end{enumerate}

\item After comprehensive summary, write the speaker summary by
    \begin{enumerate}
        \item Prior to writing the speaker summary, identify the key speakers based on the comprehensive summary. 
            \begin{itemize}
                \item Usually whichever speakers spoke the most, but also consider those contributed to the \textit{key moments}
            \end{itemize}
        \item For each key speaker, reread only their comments in the original conversation. Then in one sentence, describe their specific changes in tone/stance/conversation strategies and interactions/responses to other key speakers
    \end{enumerate}

\end{enumerate}

\subsection{Interactive Work}
Annotators start the interaction from the following setup:
\begin{itemize}
    \item Annotator A: having completed the individual work for the conversation, i.e., read the original conversation and wrote the comprehensive summary and the speaker summary
    \item Annotator B: didn’t read the original conversation, now writes the summary of conversation dynamics. 
\end{itemize}

Collaboratively, they follow these steps, which we describe from a third-person perspective for better clarity.

\begin{enumerate}
    \item Annotator B reads the comprehensive summary and speaker summary out loud. They ask initial questions to Annotator A confirming the order of speaker comments and key speakers (“Speaker1 then Speaker2 then Speaker1 again?”, “Speaker1 spoke the most?”), the overall stance/speaker relationship of the argument (“Speaker1 and 3 agreed, and both disagreed with Speaker2?”)

    \item  Annotator B begins writing the \SCD by first copying the first sentence of the comprehensive summary, which often describes the overall topic of the conversation in a few words. 
    \item Annotator B identifies the first \textit{section} of the comprehensive summary, highlighting the summary sentences on the document so that Annotator A can also reference. 
    \begin{itemize}
        \item \textit{section} – usually 1-3 \textit{comments} that fall before/in between any \textit{key moments}. These \textit{comments} should have a similar impact on the overall conversation dynamics, so that it makes sense to condense them into one sentence in the \SCD
        \item Annotator A may disagree with condensing the section if they think important information from within the section would be lost (e.g. different tone/rhetorical elements, argumentative stances)
    \end{itemize}
    \item For each \textit{section}, Annotator B writes a corresponding summary capturing the dialogue acts, conversation strategies, and tonal elements, without any topical details. 
    \begin{itemize}
        \item dialogue acts and conversation strategies examples: disagreement, agreement, counterargument, criticism, accusation, providing sources, requesting sources, insulting, defending, acknowledging, conceding, rhetorical questions, invalidating, repetition, using long comment
        \item tonal elements example: sarcasm, passive-aggressiveness, bluntness, rudeness, civility, neutrality, passion, harshness, strength, assertiveness, politeness, friendliness, objectivity, annoyance, frustration, tension, provocation, skepticism, demanding
        \item If the indicator of tone is missing or not clear, Annotator B asks Annotator A questions such as the ones below, and Annotator A often goes back to the original conversation to reread comments and provide accurate answers to the questions or even read aloud whole phrases of a comment if needed to give proper context
        \begin{itemize}
            \item B: “Was this said neutrally, or is there something about the tone that I should note?”
            \item B: “Is the comment overtly rude, or is it just passive-aggressive or blunt?”
        \end{itemize}

    \item  Annotator A reviews the work done on this \textit{section} and makes corrections or suggestions if they think the conversation dynamics summary isn’t an accurate representation of the conversation. And, Annotator A and B would revise the sentences together. 

    \item They repeat this process for each \textit{section}.
    \item Annotator A rereads the whole conversation dynamics summary, noting if any part does not seem to accurately reflect the original conversation/comprehensive summary. Both people work together to correct any such cases with the question-asking method above. 
    \begin{itemize}
        \item If needed, annotators would condense the summaries to be under 80 words, but usually they were already within range.
    \end{itemize}

    \end{itemize}

\end{enumerate}

\section{Informativeness Check}
\label{append:informativeness}
\xhdr{Conversations covered in the check}
We first sample 10 conversations on 10 different topics. 5 of the conversations are `derailing' and 5 are `non-derailing'. Each of these conversations makes one question, where this conversation offers its transcript and a segment from its summary as the correct choice. The paired conversations of these 10 conversations offer the first type of distractors as discussed in the main text (same topic but opposite derailment label). Then, for the second distractor of each question, we use a conversation that has a different topic but the same derailment label as the correct choice. 
We also ensure that each conversation is used only once across all questions (either offering a transcript and correct choice or offering a distractor). 
This way, each choice in the question represents a unique conversation and we maximize the coverage of our check, covering a total of 30 conversations. 

\xhdr{Extracting and processing segments}
For this basic check, we define a segment as a sentence that has 2 speakers. Each summary would have multiple segments and we always randomly select one. For the three segments (choices) of a question, we rename them in such a way that the speakers in all three segments have the same pseudonames. The speakers in the transcript is also renamed accordingly to be consistent with the correct segment.  This effort prevents a question from being trivial when, for example, ``Speaker5'' appears in a distractor but never appears in the transcript, which immediately rules out this distractor. With this renaming, annotators have to carefully read all 3 choices against the transcript to answer a question correctly.\footnote{Before this renaming, we've already anonymized the speakers' usernames with Speaker1, Speaker2, etc., to respect the their identity.}

\section{Human Forecasting Experiment}
\label{appendix:eval}                                                                                                       
We now discuss our design for evaluating human forecasting on conversation summaries.
To design experiments that respect the annotators' attention span, we divide the 20 conversations into two batches of 10 conversations for 2 rounds of exercises with the same procedure. All annotators participate in both rounds. 

In each round, we have 10 subjects divided into 2 groups (A and B), each completing one version of our questionnaires (each containing 10 summaries). The questions are designed such that the $i$-th question in either questionnaire presents a summary for the same conversation but the summaries are created differently (one is human-written and the other is machine-generated). For example, if the $i$-th question in Questionnaire A is a human-written summary for a conversation, then the $i$-th question in Questionnaire B is a \gptelaborate summary for the same conversation. 
This way, each participant has an equal weight on the results for human-written and machine-generated summaries, and thus any difference between these results can not be attributed to a single annotator (e.g., that is exceptionally good at the forecasting task).

For each conversation, the annotator sees the conversation summary and is asked to guess whether the conversation will derail in the future and give scores for their confidence in their guess, topic understanding, and conversation trajectory understanding. 
We briefly define conversation trajectory at the start of the questionnaire, as how the interaction between speakers evolves during the discussion, independent of the actual topics discussed. 
Additionally, we also record the time between the subjects seeing the summary and submitting the forecast.
Figure \ref{fig:exercise1} presents an example question. 
After the experiment, we also debriefed the subjects to understand how they understood the questions; one observation that stood out was confusion regarding the trajectory scale and how that relates to the guess they are making.

For evaluating human forecasting on transcripts, we follow a similar design with some modifications. First, we have a different group of 10 participants, such that there is no pollution between the two experiments.  
Since reading a transcript requires much longer time than reading a summary, each participant only reads 10 transcripts, with the exception of 2 participants who volunteered to read all 20 transcripts.  This results in 6 guesses for each transcript. 

\begin{figure*}[ht]
    \centering
    \fbox{
        \parbox{\dimexpr\textwidth-2\fboxsep-2\fboxrule\relax}{
            \textbf{[Conversation Summary]} \\
Speakers discuss the responsibilities of caregivers of autistic children. One Speaker opens up the discussion using strong language. Speaker3 and Speaker4 begin to argue in a passive-aggressive manner, which then transitions into sarcasm, accusations, and questioning each other's logic. Speaker4 supports their point with a personal experience, which Speaker3 refutes rudely.
\newline
\textbf{Will the conversation go awry (derail)?}
  \begin{itemize}
    \item[\ocircle] Yes
    \item[\ocircle] No

  \end{itemize}

        }
    } 
    \vspace{10pt} 
    \fbox{
        \parbox{\dimexpr\textwidth-2\fboxsep-2\fboxrule\relax}{
  
 \textbf{Confidence of your answer (1 for least confident and 5 for most confident)}
  \begin{itemize}
    \item[\ocircle] 1
    \item[\ocircle] 2
    \item[\ocircle] 3
    \item[\ocircle] 4
    \item[\ocircle] 5
  \end{itemize}

\textbf{To what extent did the summary help you understand the topic of the conversation (on a scale of 1 to 5)?}
  \begin{itemize}
    \item[\ocircle] 1: I don't even know the general topic.
    \item[\ocircle] 2
    \item[\ocircle] 3: I know the general topic of the discussion.
    \item[\ocircle] 4
    \item[\ocircle] 5: I know how each Speaker is related to the topic.
  \end{itemize}

  \textbf{To what extent did the summary help you understand the conversation trajectory (on a scale of 1 to 5)?}
  \begin{itemize}
    \item[\ocircle] 1: I don't have any idea of the trajectory of the conversation.
    \item[\ocircle] 2
    \item[\ocircle] 3: I have a general understanding of the trajectory.
    \item[\ocircle] 4
    \item[\ocircle] 5: I have a thorough understanding of how each Speaker contributed to the trajectory.
  \end{itemize}

        }
    }
    \caption{Example question for derailment forecasting based on summaries.}
    \label{fig:exercise1}
\end{figure*}

\section{Details of Summarization Models}
\label{appendix:autosumm}
\subsection{Generating Multiple Summaries For a Conversation}
\label{append:4trials}
For every summary type (e.g., traditional prompt, procedural prompt, finetuned BART), we repeat the process of generating summaries and running the downstream evaluation in 4 trials, each trial generating a different summary for a conversation. For a summary type based on a finetuned model, in each trial we finetune the model with a different random seed for summary generation. For a summary type based on prompting GPT-3.5-turbo, we simply utilizes its stochasticity, using its default sampling parameters to generate a new summary for each conversation.

\subsection{Finetuned Summarization Systems}
For finetuned summarization systems, we use 40 transcript-summary pairs from our human summary dataset for finetuning, 10 pairs for development, and generate summaries for the remaining 50 test set conversations that do not have human summaries. The generated summaries are then evaluated with our downstream task in Section \ref{sec:downstream}. 

We first experimented with the SOTA conversation summarization systems, BART-large and DialogLED \citep{lewis_bart_2020, zhong_dialoglm_2022}. Both systems previously showed strong performance on long dialogue summarization datasets with small train sets, such as AMI (train size 97) \citep{carletta_ami_2006} and ICSI (train size 43) \citep{janin_icsi_2003}, as reported in \citet{zhong_dialoglm_2022}.
Table \ref{table:finetuned} reports the performance brought by summaries from finetuned BART and DialogLED in our downstream task. We find that these models finetuned on the 40 human written summaries, do not produce summaries that lead to better forecasting results than \gptelaborate summaries. 

Additionally, we attempted to finetune GPT-3.5-turbo using OpenAI's API. Due to the high cost OpenAI charges for finetuning and inferencing on finetuned checkpoints, we find adequate hyperparameter search unfeasible and stopped after obtaining one checkpoint with reasonable summary quality. The summaries by this checkpoint led to an accuracy of 61.9\% in the downstream task, substantially lower than the accuracy brought by the procedural prompt summaries (Table \ref{table:finetuned}).

\begin{table}[t]
\centering
\begin{tabular}{lc}
\hline
 \textbf{Based on...} & \textbf{Accuracy} \\ 
transcripts (subset) & 56.0  \\ 
procedu. prompt summ. (subset) & \textbf{71.5} (2.5)  \\ \hline  \hline
BART summ. (subset) & 57.5 (3.0) \\ 
DialogLED summ. (subset) & 55.0 (4.2) \\ 
\bottomrule

\end{tabular}
\caption{Few-shot GPT derailment forecaster performance based on finetuned models summaries (for the 50 test set conversations that do not have human summaries). We include results on transcripts and procedural prompt summaries of the same 50 conversations for reference.} 
\label{table:finetuned}
\end{table}

\subsection{Other Forecasting Systems}
\label{append:other_cls}
For using GPT-3.5-turbo as few-shot classifiers, we set the sampling temperature to 0 for deterministic behaviors. 
\newline

Additionally, we also experimented with other classifiers using supervised training to forecast conversation derailment. We use the transcripts or the generated summaries of the train (234 conversations) and dev (100 conversations) splits of our dataset to obtain trained classifiers and run inference on the transcripts or generated summaries of the test split (100 conversations). We examine two strong baseline models for text classification for this supervised setting, namely BART and Longformer. Although these supervised models are consistently outperformed by the GPT few-shot classifier (Table \ref{table:compare_cls}), when comparing their performances on the generated summaries, we still find that \gptelaborate summaries best help the downstream forecasting of conversation derailment, indicating that our conversation \subtext summary task indeed helps automatic systems to forecast conversation derailment.

\subsection{Prompt Engineering}
When developing our zeroshot and procedural prompts for \subtext summaries, we tried different synonyms for conversation \subtext and specific \subtext elements, as well as changing the phrasing of their definitions and examples. For example, instead of simply prompting the model to summarize `conversation \subtext', which might appear as a novel jargon to the model's parametric knowledge, we instruct the model to write a summary that captures the trajectory of the conversation, especially focusing on how elements like tone, sentiment, conversation strategies may change or remain the same throughout the conversation. We then manually examine the quality of generated summaries for a small prompt engineering dataset (size 10) that's disjoint with our dev and test splits. 
For the procedural prompts, in particular, we manually wrote example summary segments to contrast different aspects of traditional summaries with those of \SCDs, and included these examples in the procedural prompt. Figure \ref{fig:prompt_contrast} shows the two prompts we eventually chose as the zeroshot and procedural prompts for \SCDs. 
\newline

\label{append:prompt_enginner}
\begin{figure*}
    \centering
    \fbox{%
        \parbox{\dimexpr\textwidth-2\fboxsep-2\fboxrule\relax}{%
            \textbf{Zeroshot Prompt:} \\
             Write a short summary capturing the trajectory of an online conversation. Do not include specific topics, claims, or arguments from the conversation. Instead, try to capture how the speakers' sentiments, intentions, and conversational/persuasive strategies change or persist throughout the conversation. Limit the trajectory summary to 80 words.
        }
    }
    \vspace{10pt} %
    \fbox{%
        \parbox{\dimexpr\textwidth-2\fboxsep-2\fboxrule\relax}{%
            \textbf{Procedural Prompt}: \\
            Write a short summary capturing the trajectory of an online conversation.  
Do not include specific topics, claims, or arguments from the conversation. The style you should avoid:

Example Sentence 1: “Speaker1, who is Asian, defended Asians and pointed out that a study found that whites, Hispanics, and blacks were accepted into universities in that order, with Asians being accepted the least. Speaker2 acknowledged that Asians have high household income, but argued that this could be a plausible explanation for the study's findings. Speaker1 disagreed and stated that the study did not take wealth into consideration.” 
This style mentions specific claims and topics, which are not needed.

Instead, do include indicators of sentiments (e.g., sarcasm, passive-aggressive, polite, frustration, attack, blame), individual intentions (e.g., agreement, disagreement, persistent-agreement, persistent-disagreement, rebuttal, defense, concession, confusion, clarification, neutral, accusation) and conversational strategies (if any) such as 'rhetorical questions', 'straw man fallacy', 'identify fallacies',  and 'appealing to emotions.' 
The following sentences demonstrate the style you should follow:

Example Sentence 2: “Both speakers have differing opinions and appeared defensive. Speaker1 attacks Speaker2 by diminishing the importance of his argument and Speaker2 blames Speaker1 for using profane words. Both speakers accuse each other of being overly judgemental of their personal qualities rather than arguments.”

Example Sentence 3: “The two speakers refuted each other with back and forth accusations. Throughout the conversation, they kept harshly fault-finding with overly critical viewpoints, creating an intense and inefficient discussion.”

Example Sentence 4: “Speaker1 attacks Speaker2 by questioning the relevance of his premise and Speaker2 blames Speaker1 for using profane words. Both speakers accuse each other of being overly judgemental of their personal qualities rather than arguments.”

Overall, the trajectory summary should capture the key moments where the tension of the conversation notably changes. Here is an example of a complete trajectory summary. 

Trajectory summary: 

Multiple users discuss minimum wage. Four speakers express their different points of view subsequently, building off of each other’s arguments. Speaker1 disagrees with a specific point from Speaker2’s argument, triggering Speaker2 to contradict Speaker1 in response. Then, Speaker3 jumps into the conversation to support Speaker1’s argument, which leads Speaker2 to adamantly defend their argument. Speaker2 then quotes a deleted comment, giving an extensive counterargument. The overall tone remains civil.

Now, provide the trajectory summary for the following conversation.

Conversation Transcript: [...]

        }
    }
    \caption{Zero-shot prompt and procedural prompt for SCDs.}
    \label{fig:prompt_contrast}
\end{figure*}

\section{Qualitative Analysis}

\label{appendix:qual_analysis}
Inspired by prior literature \citep{habernal_before_2018, zeng_what_2020, zhang_conversations_2018, li_exploring_2020}, we focus on a set of \strategies related to conversation trajectories for our qualitative analysis in Section \ref{sec:eval}. Here, we present the list in Table \ref{table:strategies}.

\section{Miscellaneous}
\label{appendix:miscellaneous}
\subsection{Transcript of the Introductory Example}
We provide the transcript of our introductory example in Figure \ref{fig:transcript1} to \ref{fig:transcript4}. The last 3 utterances of the transcript are omitted as how it appears in our dataset.

\subsection{Data Collection}
\xhdr{Annonymization} We collect human summaries for conversation transcripts from the published dataset CGA, which we accessed through ConvoKit 2.5.3. The dataset contains the usernames of the conversation participants, which we replace with `Speaker1', `Speaker2', and etc. to respect the users' identity. 

\xhdr{Annotators} All annotators for our evaluations are recruited as volunteers from university students in the US. The two annotators who wrote the summaries of conversation dynamics are co-authors of this paper. The data collection was approved by an Institutional Review Board at the authors' institution. All annotators were informed that their data would be used for an NLP research and eventually a published paper before they gave consent. 

\xhdr{Disclaimer of Risks} All annotators are informed that ``some of the conversations presented in the annotation task can be extremely biased and offensive and speak of sensitive topics.'' All annotators gave their consent to participate.

\subsection{Implementation Details}
For our finetuned models, we conducted hyperparameter search over learning rates [3e-6, 5e-6, 1e-5, 2e-5, 3e-5, 5e-5, 1e-4] and warmup steps ([40, 80] for summarizers and [234, 468] for classifiers), and used the default values from their original implementation for other hyperparameters. For the DialogLED and BART summarizers, we eventually used a learning rate of 3e-5 and 80 warmup steps. For the BART classifier, we used a learning rate of 3e-6 and 468 warmup steps. For the Longformer classifier, we used a learning rate of 5e-6 and 468 warmup steps. The finetuning experiments in total took about 150 GPU hrs on an Nvidia A40 GPU.

\subsection{Used Artifacts}
We include a list of existing artifacts we used. Some of them have been cited in the main sections of this paper above. We have closely followed their intended use. 
\begin{itemize}
    \item GPT-3.5-turbo-0613:\\ a snapshot of GPT-3.5-turbo from June 13th, 2023. Closed-source but accessible at a low cost via OpenAI's API, https://platform.openai.com/docs/
    \item ConvoKit 2.5.3:\\ https://convokit.cornell.edu/, MIT License
    \item PyTorch 1.8:\\ https://pytorch.org, BSD-3 License
    \item Transformers 4.25:\\ https://github.com/huggingface/transformers, Apache License 2.0
    \item Scikit-learn 1.3.2:\\ https://scikit-learn.org, BSD-3 License
    
\end{itemize}

\subsection{Additional Evaluation Metrics}
\label{subsection:additional_metrics}
Here, we provide additional performance metrics (precision, recall, macro-averaged F1) for different summary types, when they are evaluated with our derailment forecasting task. Each summary type is evaluated with its respective GPT-3.5 few-shot derailment forecasting model as described in Section~\ref{subsection:forecaster}. 

\begin{table}[H]
\centering
\begin{tabular}{llll}
\toprule
Derailing?             & prec. & rec. & F1 \\
\midrule
       False &    72.7 &   32.0 &   44.4 \\
      True &    56.4 &   88.0 &   68.8 \\
\midrule
   macro avg &    64.6 &    60.0 &    56.6 \\
\bottomrule
\end{tabular}
\caption{Additional metrics for derailment forecasting on \textbf{transcripts}}
\end{table}

\begin{table}[H]
\centering
\begin{tabular}{llll}
\toprule
Derailing?             & prec. & rec. & F1 \\
\midrule
       False &   57.1  &  66.5  &  61.4  \\
      True &   59.9  &  50.0  & 54.5   \\
\midrule
   macro avg & 58.5    & 58.3    &  58.0  \\
\bottomrule
\end{tabular}
\caption{Additional metrics for derailment forecasting on \textbf{traditional summaries}}
\end{table}

\begin{table}[H]
\centering
\begin{tabular}{llll}
\toprule
Derailing?             & prec. & rec. & F1 \\
\midrule
       False &  56.1  &  80.0  & 66.0   \\
      True & 65.2   & 37.5   & 47.6  \\
\midrule
   macro avg &  60.7   & 58.8    & 56.8  \\
\bottomrule
\end{tabular}
\caption{Additional metrics for derailment forecasting on \textbf{zeroshot prompt summaries}}
\end{table}

\begin{table}[H]
\centering
\begin{tabular}{llll}
\toprule
Derailing?             & prec. & rec. & F1 \\
\midrule
       False &  62.9  &  84.0  & 72.0   \\
      True &  75.9  &  50.5  & 60.7  \\
\midrule
   macro avg &  69.4   &  67.3   &  66.3 \\
\bottomrule
\end{tabular}
\caption{Additional metrics for derailment forecasting on \textbf{procedural prompt summaries}}
\end{table}

\begin{table}[H]
\centering
\begin{tabular}{llll}
\toprule
Derailing?             & prec. & rec. & F1 \\
\midrule
       False & 57.6   &  65.0  &  60.6  \\
      True &  59.2  & 51.0   & 54.2   \\
\midrule
   macro avg & 58.4    &   58.0  & 57.4  \\
\bottomrule
\end{tabular}
\caption{Additional metrics for derailment forecasting on \textbf{finetuned BART summaries}}
\end{table}

\begin{table}[H]
\centering
\begin{tabular}{llll}
\toprule
Derailing?             & prec. & rec. & F1 \\
\midrule
       False & 56.1   &  49.0  &   50.3 \\
      True & 54.4   &  60.0  & 55.6 \\
\midrule
   macro avg & 55.3   & 54.5    &52.9   \\
\bottomrule
\end{tabular}
\caption{Additional metrics for derailment forecasting on \textbf{finetuned DialogLED summaries}}
\end{table}

\clearpage
\begin{table*}[h!]
\centering
\begin{tabular}{@{}p{0.35\textwidth}p{0.6\textwidth}@{}}
\toprule
\textbf{Strategies}   & \textbf{How they can be mentioned in \subtext summaries} \\ \midrule
Rhetorical questions &  ``poses a rheotrical question'', ``rhetorically asks'' \\ \hline
Attacking logic &   ``point out flaws in [the other speaker]'s arguments'', ``accuses [the other speaker] of their logical fallacy''\\ \hline
Anecdotal experience & ``shares a personal story'', ``uses an anecdotal example'' \\ \hline
Evidence & ``cites statistics and data to support their viewpoint'', `uses external sources to support'' \\ \hline
Juxtaposition & ``makes a comparison between'', ``provides a detailed explanation of the differences between'' \\ \hline
Analogy & ``uses an analogy to support''\\ \hline
Pointing at missing or unsupported evidence & ``asks for evidence'', ``criticizes the lack of evidence'' \\ \hline
Accusing of not correctly treating their argument &  ``accuses [the other speaker] of not reading their arguments'', ``accuses [the other speaker] of reinterpreting their positions'' \\ \hline
Questioning one's knowledge or attacking one's lack of knowledge & ``insulting [the other speaker]'s knowledge of [the subject]'', ``accusing [the other speaker] of lacking the knowledge of [the subject]''\\ \hline
Hypothetical example & ``proposing another hypothetical scenario'' \\ \hline
Counterexample & ``presents counterexamples''\\ 
\bottomrule
\end{tabular}
\caption{List of \strategies a speaker may use. For our qualitative analysis, we consider an \SCD `mentions' a \strategy only if it explicitly identifies what the strategy is, as shown in the examples listed in the second column of this table. For example, if a summary simply paraphrases the exact rhetorical question or the cited evidence, then we do not count it as `mentioning' a \strategy.} 
\label{table:strategies}
\end{table*}

\clearpage
\begin{figure*}[t!]

    \centering
    \fbox{
        \begin{minipage}{0.9\textwidth}
\textbf{Transcript}: 

Speaker1: Businesses aren't charities. They exist to make a profit. "Morals" and "ethics" are always trumped by profit  in the business world. 

Speaker2: That's.... Kind of the inherent problem.

Speaker3: For you. Whenever there is a "problem", it is usually some party wanting to further their interests. Remember that morals do not exist out there, they are a construct of society. If a majority is disadvantaged, they may use "morals" to push for their interests. 

Speaker2: Unchecked capitalism is economically unsustainable. So yes, it is a problem for me, and for everyone participating in the economy. Business involves more than just profits. It involves human beings investing their labor. These are not machines. The idea that profit alone should drive our economic decisions is morally bankrupt. Human beings deserve a modicum of dignity. If you can't agree to that, and you think that slavery is OK and justified as long as the business is profitable, then I would posit that you too are morally bankrupt.

Speaker4: "Unchecked capitalism is economically unsustainable." what do you mean "unsustainable"?  what happens?

edit: [MFW I get my daily reddit downvotes for questioning a socialist](http://i.imgur.com/QoGM3.gif)

Speaker2: Well, let's have an economic thought experiment. 

Rule 1: Businesses only care about profit (and typically the short term profit at that, not longer term returns).

Rule 2: Businesses have nothing to check them from abusive practices.

Let's think of some common things people believe are good about this scenario:

1) Everyone seeks to further their own means so with supply and demand everything works out in the end!

2) If a company has poor practices, the consumer will migrate to other options.

3) If a company has poor practices, another company will be developed to compete against it, forcing it to rid itself of those poor practices.

Now let's think about whether or not those thoughts can sustain themselves in the thought experiment.

1) Why wouldn't companies seek to form cartels? We already have evidence that they do this, even among competitors. You might say that they'll be seeking to further their own means, so this is only temporary. But I would ask you to consider that if Company A and Company B can make larger profit margins colluding than they can competing, why wouldn't they? Sure, Company A might be able to make slightly higher profits down the road if they were able to beat Company B outright, but that takes short term investment (and therefore cost) and effort. With collusion, we can maximize profits with minimal effort and cost. 

2) But surely if A\&B collude with one another, C will come out of the woodwork and offer a better product/service at a better price, right? But will they? If A\&B catch wind of C, what's to stop them from using potentially coercive means from stopping them from competing. For example: hostile takeovers. A\&B are a cartel now, and they are established, so they certainly have the capital necessary to absorb C as a fledgling company. 

3) If A\&B are colluding, what's to stop them from slowly degrading the conditions of their workforce? Like the employees of office space only working hard enough not to get fired, A\&B are only going to create conditions good enough as to not lose profit margins. 

4) If A\&B are able to block competitors from the industry, what motivation do they have to innovate and improve their products? 

5) What's to stop A\&B from doing environmentally disastrous things (pollution for example)? If short term profits are all that matter, why would they bother to care?

[continues on the next page..]
        \end{minipage}
        
    }
    \caption{Transcript of the conversation used for the introductory example, Part 1 of 4. Its conversation ID is cz2r8ig in the CGA-CMV corpus.}
    \label{fig:transcript1}
\end{figure*}

\clearpage
\begin{figure*}[t!]
    \centering
    \fbox{
      \begin{minipage}{0.9\textwidth}
[continues from the previous page] \\
All of this isn't to say that Government must be the check on businesses. Government can be the tool of business to enshrine their power as well. But capitalism by itself can have some pretty negative consequences when it is extrapolated out. The problem is the fundamentals of capitalism are built on a foundation of assumption. That assumption is that when people seek to fulfill their individual desires, with supply and demand and perfect competition at play, that everything reaches an equilibrium. The problem here is that perfect competition is never really at play. Like most things "perfect" it's an idea, not a reality. You can't enforce perfect competition. So the next closest thing you can do is create an environment where competition is maintained and our perpetually imperfect competition doesn't get too out of control. 

Personally, I'm a believer in social democracy - I believe that capitalism can be good when it is juxtaposed by the idea of solidarity with all of mankind (as opposed to purely selfish means). That's not communism, mind you, it just means that we shouldn't lose sight of humanity in the process. When we think about that as a counter to selfish greed, we start thinking about longer term returns, we start thinking about societal benefit, we start humanizing the capitalistic process (they aren't "capital" or "human resources" or "employees" they're people investing their efforts in making these companies perform). 

I hope that addresses some of your question. It was a bit open ended, so I tried not to ramble on too much. 

Speaker4: that addressed ZERO of my question. you completely failed to address what happens when a capitalist system reaches its alleged "sustainability" threshold.

my question was not open-ended.  what occurs when it is no longer sustainable? you said it was unsustainable. your hypothetical "what ifs" don't support your assertion.

Speaker2: Ok. Thanks for the attitude. Now I'm just going to give you a curt response.

Economic collapse. Mass unemployment. Overly polluted and toxic environments. Did you just want a parade of horribles? 

Speaker4: attitude? i just explained that you didn't answer my question.  you gave me an opinionated rant instead of an explanation.

what am i supposed to do, just say "oh thanks"?

"economic collapse"
yeah, you said that already. why? how?

"mass unemployment"
how? everybody just gradually becomes unemployed "because capitalism"?

"toxic environments"
why? is there some aspect of socialism that prevents toxicity in products?  does socialism provide some sort of waste-disposal service unavailable in a capitalist system?

you just keep throwing out matter-of-fact assertions, but i don't see how you are arriving at your conclusions.

and apparently you interpret scrutiny as "attitude" that you take offense to... i dunno.  I'm not convinced.

edit: annnnd im downvoted instead of having any of those valid questions answered.

Speaker2: "attitude? i just explained that you didn't answer my question. you gave me an opinionated rant instead of an explanation."
Sorry, perhaps I'm just reading into the emphasis from "ZERO" and "completely". The tone of your response and the one that followed seemed to be ... dickish, for lack of a better word. If I'm reading into it, my apologies.

Economic collapse. Increasingly volatile market behaviors as a result of increasingly risky investments. Essentially: short term profit at any cost. The system fails to account for long term sustainability. 

[continues on the next page..]
        \end{minipage}
        
    }
    \caption{Transcript of the conversation used for the introductory example, Part 2 of 4.}
    \label{fig:transcript2}
\end{figure*}

\clearpage
\begin{figure*}[t!]
    \centering
    \fbox{
       \begin{minipage}{0.9\textwidth}
[continues from the previous page] \\
Mass unemployment. Actually, we're already on our way to mass unemployment due to automation. I just read an article about us losing 5 million (net) jobs by 2020 to automation and AI (that includes the 2mil that will pick up new jobs from the new tech). You also have an increasing population who can be provided for but not necessarily enough jobs to have them making a worthy contribution. 

Toxic environments. I actually didn't mention socialism, you did. I said social democracy. The primary economic driver of social democracy is capitalism. It's just checked. But even still: yes. There is something inherent in those systems that greatly reduces the risk of toxicity. The people who make the products and and invest the labor are the ones who realize the gains. They are also the ones to decide whether or not those gains are worthwhile given the risks. 

In a raw capitalism system: the capitalist realizes the gains and, due to the increased capital, can largely insulate themselves from the associated risks. Drinking water pollution is an example of this. The rich typically don't have this problem as they can afford to purchase potable water. The poor and working class may not always have their luxury (Flint, MI for example). So the people who have to live with the realities and consequences of their production are the ones making the decisions as opposed to someone who doesn't have to live with them or who can use their financially superior position to avoid them. 

Other examples include the short term profits associated with taking unnecessary risks: see countless oil spills and deepwater horizon. Also see: covering up of global warming by oil companies or covering up of health hazards by cigarette companies. 

The capitalist system *encourages* this behavior. It is financially beneficial (short term profits are encouraged over long term profits and investment - this is legally supported through cases dating back to Dodge v. Ford). A checked capitalist economy disincentivizes that behavior through regulation and social welfare. A socialist economic system would probably do the same through the reality that people who would be causing the harm are the people who have to live with the harm and are less likely to remove themselves from it. 

If we want to debate the core tenets of a capitalist economy: that's fine. But I operate on the presumption that we both agree the goal of a capitalist economy is to generate profit (with a weight towards short term profit especially with larger firms having to provide quarterly earnings and financial benefit for shareholders). 

To get back to OP - the minimum wage is *one* check on the default capitalist economic framework (to drive costs as low as possible). OSHA is another. The FDA is another. The SEC is another. And so on and so forth. 

Speaker4: thank you for that thoughtful response.  i do type in a very tonal, speech-based style, so people actually accuse me of being "a dick" or "irate" pretty frequently. i guess i will just have to get used to it.

i just mentioned socialism as an alternative to compare to in the pollution argument, i didn't mean to presume to say you were, necessarily.  my apologies if it seemed that way.

So anyway, yes. I agree that capitalism creates volatile markets.  However, I don't consider it to be as threatening as you seem to.  I can't understand any situation that would cause a cataclysmic death or cessation of those markets. There will always simply be peaks and valleys.

"The system fails to account for long term sustainability."

[continues on the next page..]
        \end{minipage}
    }
    \caption{Transcript of the conversation used for the introductory example, Part 3 of 4.}
    \label{fig:transcript3}

\end{figure*}

\clearpage
\begin{figure*}[t!]
    \centering
    \fbox{
      \begin{minipage}{0.9\textwidth}
[continues from the previous page] \\
This is really vague.  What happens when it no longer becomes sustainable? Anarchy?  Does the wealth stop existing? Where does it go? Does it become impossible to make purchases or be paid for your work? Does everyone die? I am really trying to pin down what exactly you mean by a system "sustaining" itself.  It seems more like a system that would have to be electively given up, like a language that dies out, instead of something that can break and has to be discarded, like a broken dish or a burnt-out light bulb.

In this sense, I don't see how any other ideology or economic system would succeed or fail in any of these areas, without the "unchecked" qualifier I see so commonly applied to critique of capitalism.

"unchecked" *is* the problem. greed is the problem. overpopulation is the problem. creation of toxic byproducts and harmful processes is the problem.  a sheltered ruling-class is the problem.  none of these things will disappear by switching ideologies.  no other economic system is better, or worse, -equipped to eradicate these things.  similarly, blaming "unchecked" application of any other system for the existence of these problems would be just as silly as blaming capitalism.

Speaker2: That was my point in the beginning though. That unchecked capitalism is bad. I am merely arguing for a check on it. Something to keep it from becoming too volatile, too disruptive, too detrimental. Something to keep it contained. I view capitalism like a nuclear reactor. If you keep the reaction going and have the right containment - you've got a nice source of clean energy. But if you fail to keep it contained - you've got bad news.

To your question of what happens. It can really mean any number of things. It could mean a very slow dystopian degradation of society wherein the rich get progressively richer and the poor become bottom feeders with short lifespans. It could mean everyone dies (perhaps through experimentation gone awry - particularly in the energy or medical fields). I'm not sure about wealth ceasing to exist outright, I don't know what that would look like or how that would come about. Anarchy is certainly possible, so is revolution (as the rich get richer, the poor get angrier - extrapolate that and we have a recipe for repeating history). 

Like I said, some of those problems can be avoided when the people profiting from the production are the people doing the production. It connects them with the consequences. It functions as a systematic check. Is it perfect? surely not. 

I agree that greed is the problem. But greed isn't a bug in capitalism, it's a feature. Like the nuclear reaction, we're trying to harness the human motivation behind greed and seize it. But, we don't want it to get too far out of control. It's a balancing act.

        \end{minipage}
    }
    \caption{Transcript of the conversation used for the introductory example, Part 4 of 4.}
    \label{fig:transcript4}

\end{figure*}

\end{document}